\documentclass{article}

\usepackage[nonatbib, final]{neurips_2020}

\usepackage[utf8]{inputenc} 
\usepackage[T1]{fontenc}    
\usepackage{hyperref}       
\usepackage{url}            
\usepackage{booktabs}       
\usepackage{amsfonts}       
\usepackage{nicefrac}       
\usepackage{microtype}      
\usepackage{amsmath,amssymb,amsthm}
\usepackage{bm}

\usepackage{microtype}
\usepackage{graphicx}
\usepackage{subcaption}
\usepackage{booktabs} 
\usepackage{wrapfig}

\usepackage[vlined,linesnumbered,ruled,norelsize]{algorithm2e}

\usepackage{pgfplots}
\pgfplotsset{compat=1.7}
\usepackage{tikz}
\usetikzlibrary{positioning,calc}

\usepackage{hyperref}

\title{Efficient Reservoir Management 
through \\ Deep Reinforcement Learning}

\author{%
Xinrun Wang\textsuperscript{\textnormal{1}},
Tarun Nair\textsuperscript{\textnormal{2}}, Haoyang Li\textsuperscript{\textnormal{1}}, Yuh Sheng Reuben Wong\textsuperscript{\textnormal{1}}, Nachiket Kelkar\textsuperscript{\textnormal{2}} \AND
Srinivas Vaidyanathan\textsuperscript{\textnormal{3}}, Rajat Nayak\textsuperscript{\textnormal{3}}, 
Bo An\textsuperscript{\textnormal{1}}, Jagdish Krishnaswamy\textsuperscript{\textnormal{2}}, Milind Tambe\textsuperscript{\textnormal{4}}
\AND 
\textsuperscript{\textnormal{1}}\textnormal{Nanyang Technological University, }\texttt{\{xwang033,liha0016,wong1109,boan\}@ntu.edu.sg}\\
\textsuperscript{\textnormal{2}}ATREE, Bangalore, India, \texttt{\{tarun.nair,nachiket.kelkar,jagdish\}@atree.org} \\
\textsuperscript{\textnormal{3}}FERAL, Bangalore, India, \texttt{\{srinivasv, rajat\}@feralindia.org} \\
\textsuperscript{\textnormal{4}}Google Research, India,
\texttt{milindtambe@google.com}
}

\begin{document}

\maketitle

\begin{abstract}
Dams impact downstream river dynamics through flow regulation and disruption of upstream-downstream linkages. However, current dam operation is far from satisfactory due to the inability to respond the complicated and uncertain dynamics of the upstream-downstream system and various usages of the reservoir. Even further, the unsatisfactory dam operation can cause floods in downstream areas. Therefore, we leverage reinforcement learning (RL) methods to compute efficient dam operation guidelines in this work. Specifically, we build offline simulators with real data and different mathematical models for the upstream inflow, i.e., generalized least square (GLS) and dynamic linear model (DLM), then use the simulator to train the state-of-the-art RL algorithms, including DDPG, TD3 and SAC. Experiments show that the simulator with DLM can efficiently model the inflow dynamics in the upstream and the dam operation policies trained by RL algorithms significantly outperform the human-generated policy.
\end{abstract}

\section{Introduction}
Dams are intended to serve multiple functions, including irrigation, hydropower, and water supply. While some dams can provide flood control functions, in the Indian context, they are largely designed to maximize storage, and operated to avoid floods following extreme storm events. While dam/reservoir operations are largely governed by project objectives and user agreements, they are also guided by contingencies such as floods. Operation policies may be based on the Standard Linear Operating Policy, Storage Zoning, Rule Curves or System Engineering Techniques such as Simulation and Optimization~\cite{Jain2019Introduction}. Dam management, however, suffers from design and operational inefficiencies, e.g., sediment trapping, non-adaptive rule curves. In addition, flow regulation and the disruption of upstream-downstream linkages due to dams also impact river dynamics and threaten aquatic ecosystems and organisms. 

While dams designed for flood control can be effective at reducing peak discharge associated with storm events and increasing discharge during dry periods, these dams also cause floods in downstream areas due to poor flood forecasting and faulty reservoir storage practices. For example, extreme rainfall and high reservoir storage combined to cause the 2018 Kerala floods (India) which resulted in the death of more than 400 people, affected 56,844.44 ha of cropped area, with economic damage exceeding USD 3 billion \cite{mishra2018kerala,GovKerala2018}. The Bansagar Dam (Madhya Pradesh, India) is another example of reservoir mismanagement, and was implicated in severe downstream flooding in 2016 and 2018 due to the practice of reaching full reservoir level (FRL) in the early part of the monsoon and the subsequent lack of a flood cushion in this period~\cite{Thakkar2018}. Ecological impacts (fish species loss and breeding disruptions in endangered species) downstream of Bansagar are also documented~\cite{joshi2014environmental,Nair2016Estimation}.

The major challenges in dam operation (i.e., reservoir management) include (a) the increasing trend of extreme rainfall events makes the prediction of rainfalls inaccurate, (b) the changes in catchment precipitation and land use cannot be immediately detected by the operator, and (c) the complicated downstream dynamics and the various water usages (e.g., ecology, agriculture, fisheries, navigation) make the operation difficult to be optimized. The efficient reservoir management requires the inflow forecasting of the upstream and the modeling of the complex dynamics in the downstream~\cite{watts2011dam}.

To address this problem, we leverage reinforcement learning (RL) approaches to explore efficient dam operation policy. Specifically, we built offline simulators with the reservoir inflow model using rainfall estimates from rain gauge and satellite observations (CHIRPS) for the reservoir catchment. As hydro-climatic systems are highly dynamic, we used the Dynamic Linear Model~(DLM)~\cite{krishnaswamy2000dynamic,krishnaswamy2015non,petris2009dynamic} in which the parameters evolve with incoming data using Kalman filter and Bayesian update of parameter distributions and the forecast  error at any time-step can be estimated and incorporated into the evolution of the model. We use this model to forecast the changes of the storage of the reservoir due to inflow from the catchment upstream due to rainfall. Experiments show that the simulator with DLM can efficiently model the inflow dynamics and the dam operation policies trained by RL algorithms significantly outperform the human-generated policy.

\section{Motivating Scenario}
\begin{wrapfigure}{r}{0.45\textwidth}
\vspace{-45pt}
\includegraphics[width=0.45\textwidth]{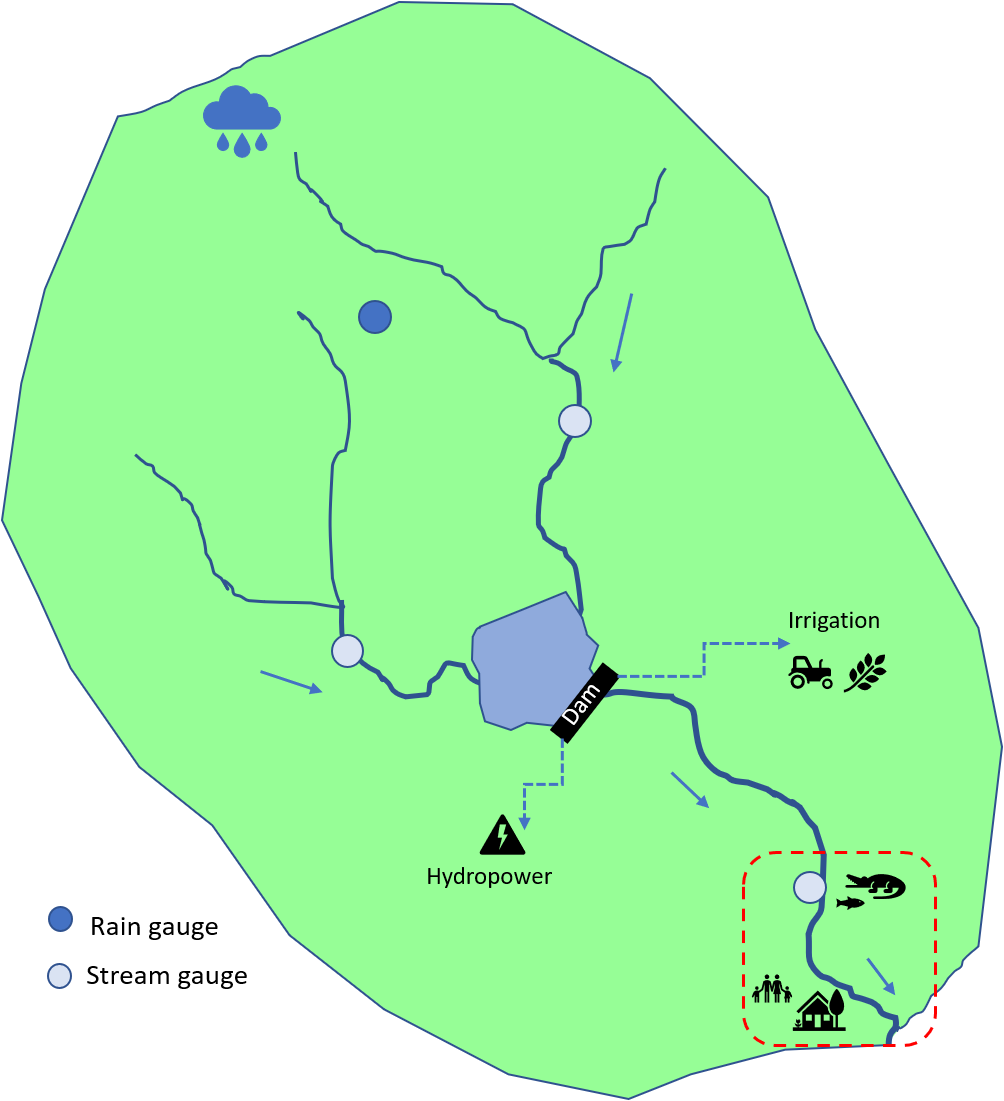}
    \caption{An example of dam management.}
    \label{fig:dam}
    \vspace{-15pt}
\end{wrapfigure}
Here we show a simple example for the dam management problem in Figure~\ref{fig:dam}, where there is a catchment area at the upstream of the reservoir and a dam which can determine the volume of water being released to the downstream. The rainfall in the reservoir catchment area will flow into the reservoir, and depends on catchment rainfall, temperature, and land use / land cover. The reservoir has two main functions: i) irrigation in the dry season, and ii) hydropower generation. The floods in the downstream can cause the death of humans and livestock, and damage crop and property. Flood damage depends on river conditions and discharge from the dam. The objective of this work is to leverage RL methods to compute the efficient dam operation policy with trained by simulators built on real data. The algorithms in this work can play a central role for the future AI-based dam operation system, which can help human operator to mitigate the downstream floods, as well as maximize various functions of the reservoir, e.g., hydropower and irrigation potentials. 


\section{Problem Statement}
The dam management problem is formulated as an MDP, denoted as a tuple $\langle\mathcal{S}, \mathcal{A}, P, R, \gamma, s_{0}\rangle$, where $\mathcal{S}$ is the set of states of the dam system, $\mathcal{A}$ is the set of actions that the dam operator can take, i.e., the amount of water being released, $P:\mathcal{S}\times\mathcal{A}\times \mathcal{S}\rightarrow [0, 1]$ is the transition function, $R:\mathcal{S}\times\mathcal{A}\rightarrow\mathbb{R}$ is the reward function, $\gamma$ is the discount factor and $s_{0}$ is the initial state. The agent's policy $\pi_{\theta}:\mathcal{S}\times\mathcal{A}\rightarrow[0,1 ]$, parameterized by $\theta$, specifies the probability of an action being taken at each state.  
The return of the agent at time step $t$ is $r_{t} = \gamma^{t}R(s_{t}, a_{t})$ where $s_{t}$, $a_{t}$ and $s_{t+1}$ are the state and the action at time step $t$, respectively. 

Specifically, the state $s\in\mathcal{S}$ consists the following information: i) the water level of the reservoir $h$ and ii) the rainfall of $K$ previous steps $\bm{rf}=\langle rf_{k}\rangle, k=1,\dots, K$ where $rf_{k}$ is the rainfall of $k$-th previous day. The transition function $P$ depends on the rainfall data at the current step $rf'$ and the dam action $a\in\mathcal{A}$: given that the current state is $s=\langle h, \bm{rf}\rangle$, the successive state $s'$ is defined 
as
\begin{align}
  h' = g(g^{-1}(h)+f(rf', \bm{rf})-a), \quad\quad \bm{rf}'=\langle rf', rf_{k}\rangle, k=1, \dots, K-1  
\end{align}
where $g$ is the function which maps the storage (i.e., the volume of water stored) to the water level of the reservoir (accordingly, $g^{-1}$ is the function mapping the water level to the storage), $f$ is the function which predict the amount of water flowing into the reservoir given the rainfall. The reward function $R$ includes three parts: i) Irrigation potentials including $R^{I}_{rice}$ which is the potential rice productivity and $R^{I}_{wheat}$ which is the potential wheat productivity, ii) Hydropower potential $R^{H}$ and iii) Flood damage $R^{F}$. The objective is to maximize the accumulated return $J(\theta)=\mathbb{E}_{\pi}[\sum_{t=0}^{\infty}r_{t}|s_{0}]=\mathbb{E}_{\pi}[\sum_{t=0}^{\infty}\gamma^{t}R(s_{t}, a_{t})|s_{0}]$.

\section{Compute Dam Operation Policy} 
We leverage the reinforcement learning methods to compute the dam operation policy. RL is a powerful method to compute the optimal policy of an MDP problem. Combining with powerful deep neural network approximators, deep RL~\cite{mnih2015human} has achieved remarkable successes on challenging discrete or continuous decision making problems~\cite{silver2017mastering}. For the problem with continuous states and/or actions, the parameterized policy $\pi_{\theta}$ can be updated by taking the derivative of the return $\nabla_{\theta}J(\theta)$. In the actor-critic method, the policy, known as actor, can be updated through the deterministic policy gradient algorithm~\cite{silver2014deterministic}:
\begin{equation}
   \nabla_{\theta}J(\theta) =\mathbb{E}_{\pi}[\nabla_{a}Q(s, a)\nabla_{\theta}\pi_{\theta}(s)]
\end{equation}
where $Q(s, a)$ is the estimated return by the critic. On the other hand, the critic can be updated through temporal difference learning, which is an update rule based on Bellman equation~\cite{sutton1988learning}. The Bellman equation describes a fundamental relationship between the value of a state-action pair $(s, a)$ and the successive $(s', a')$:
\begin{equation}
    Q(s, a) = r + \gamma \mathbb{E}_{a'\sim \pi(s')}[Q(s', a')]
\end{equation}
For large state space, the value can be estimated by a differentiable function approximator, parameterized by $\phi$. In Deep Deterministic Policy Gradient (DDPG)~\cite{lillicrap2016continuous}, both actor and critic are parameterized by networks and the critic is updated following the deep Q-learning~\cite{mnih2015human}, where the network is updated by using temporal difference learning with a secondary frozen target network $Q_{\phi'}(s, a)$ to maintain a fixed objective $y$ over multiple updates
\begin{equation}
    y=r + \gamma Q_{\phi'}(s', a'), \quad\quad a'\sim \pi_{\theta'}(s')
\end{equation}
where $a'$ is sampled from a target actor network $\pi_{\theta'}$. To reduce the overestimation of Q values in DDPG, Twin Delayed Deep Deterministic policy gradient algorithm (TD3)~\cite{fujimoto2018addressing} uses two Q-functions instead of one (hence ``twin''), and uses the smaller of the two Q-values to form the targets in the Bellman error loss functions and updates the policy (and target networks) less frequently than the Q-function. On the other hand, Soft Actor-Critic~(SAC)~\cite{haarnoja2018soft} maximizes a trade-off between expected return and entropy where the entropy is a measure of randomness in the policy, resulting the improvement of the stability during training. 

\section{Experimental Results}
\begin{wrapfigure}{r}{0.35\textwidth}
\vspace{-30pt}
\includegraphics[width=0.35\textwidth]{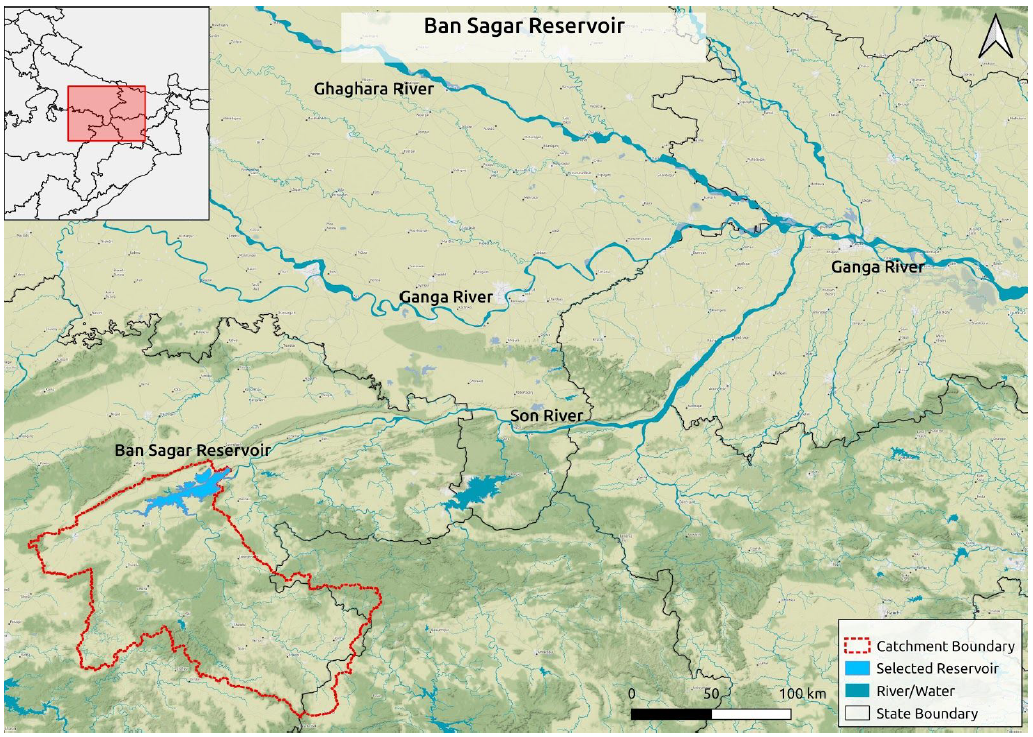}
    \caption{Bansagar dam}
    \label{fig:bansaga}
    \vspace{-15pt}
\end{wrapfigure}
\paragraph{Bansagar Dam Dataset.} An overview of Bansagar Dam's landscape is displayed in Figure~\ref{fig:bansaga}. The dataset contains data on Bansagar Reservoir water level and Rainfall received in the upper catchment area from 1 Jan 2012 to 12 Dec 2019. 

\paragraph{Simulators.} We build four simulators with different upstream models, i.e., $f(\cdot)$: i) $f(\cdot)$ is a
Generalized Least Squares~(GLS) with the rainfall data, ii) $f(\cdot)$ is a Dynamic Linear Model~(DLM) with the rainfall data, iii) $f(\cdot)$ is a Dynamic Linear Model~(DLM) with the rainfall data and the output of the GLS model and iv) $f(\cdot)$ is computed by the real data. For the other functions, we have:
i) Storage-water level model $g^{-1}:= 0.3653 h - 119.78$ where $h$ is the water level,
ii) Flood damage model $R^{F}:=e^{-981}\cdot \hat{h}^{170}$ where $\hat{h}$ is the maximum reservoir level observed during the 14 days period
iii) Irrigation potentials with a) potential rice productivity $R^{I}_{rice} = 0.1315 h-43.121$ and b) potential wheat productivity $R^{I}_{wheat} = 0.2642h - 86.641$ where the outputs are million tonnes
and iv) Hydropower potential $R^{H} = 5.1927 h - 1342.5$ where the output is million watts. Other parameters of the simulator is displayed in Table~\ref{tab:parameters}.


\begin{table}[ht]
\centering
\caption{Simulator's parameters}
\label{tab:parameters}
\begin{tabular}{ll|ll}
\toprule
 dam\_cap & 342.934 & dam\_break\_damage & 80\\
 dam\_base\_water & 0.1 &  water\_year\_start & 01 June\\
 water\_year\_end &  31 May & dry\_season\_months & Nov-Jun \\
 discount factor & 0.999 & max\_step & 365\\
 flooded\_area\_slope & 0.00006 & power\_potential\_slope & 0.003\\
 wheat\_slope & 30 &  rice\_slope & 30\\ 
\bottomrule
\end{tabular}
\end{table}

\begin{wraptable}{r}{0.45\textwidth}
\centering
\vspace{-5pt}
\caption{NSE of Different inflow model}
\label{tab:nse}
\begin{tabular}{cccc}
\toprule
     & GLS & DLM & GLS+DLM\\\midrule
  NSE   & 0.4045& 0.9843 & 0.9843\\
\bottomrule
\end{tabular}
\vspace{-5pt}
\end{wraptable}
\paragraph{Evaluation.} For the RL algorithms, we choose the widely used DDPG and two state-of-the-art methods TD3 and SAC. We first build all the four simulators with the data from 2012 to 2018, which is used for training the RL policies, and test the trained policies on the simulator where $f(\cdot)$ is computed with the data of 2019. We compare the performances of the computed polices with a human-generated discharge policy, which is displayed in Appendix~A. 

\begin{figure}[ht]
\vspace{-9pt}
\centering
\begin{subfigure}[b]{0.245\textwidth}
\includegraphics[width=\linewidth]{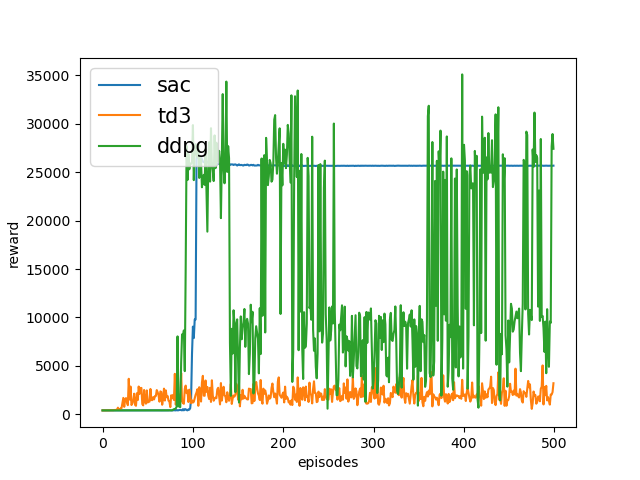}
\subcaption{GLS}
\label{fig:gls}
\end{subfigure}
\begin{subfigure}[b]{0.245\textwidth}
\includegraphics[width=\linewidth]{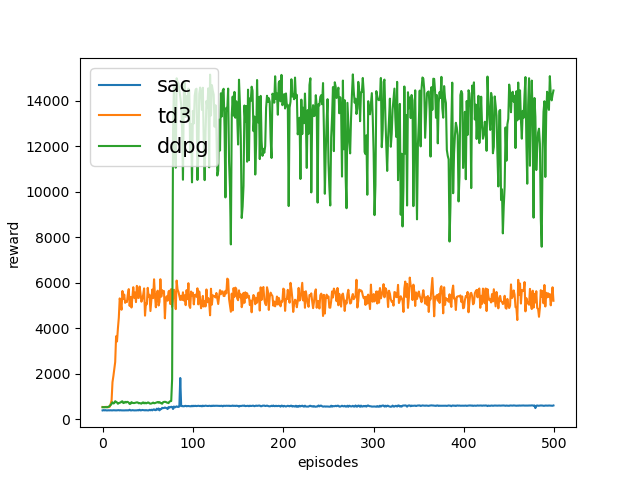}
\subcaption{DLM}
\label{fig:dlm}
\end{subfigure}
\begin{subfigure}[b]{0.245\textwidth}
\includegraphics[width=\linewidth]{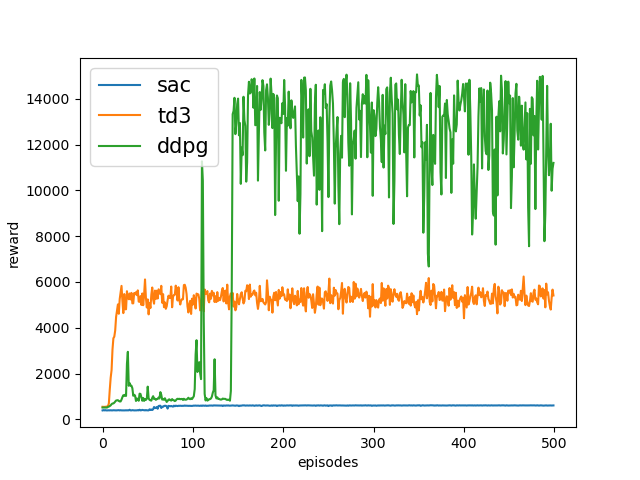}
\subcaption{GLS+DLM}
\label{fig:gls_dlm}
\end{subfigure}
\begin{subfigure}[b]{0.245\textwidth}
\includegraphics[width=\linewidth]{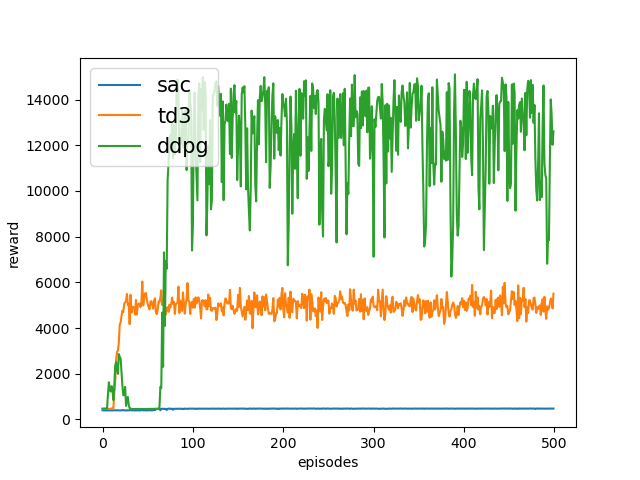}
\subcaption{Real data}
\label{fig:realdata}
\end{subfigure}
\caption{Training curves of DDPG, TD3 and SAC in different simulators}
\label{fig:training}
\vspace{-15pt}
\end{figure}

\begin{wrapfigure}{r}{0.4\textwidth}
\vspace{-15pt}
\includegraphics[width=0.4\textwidth]{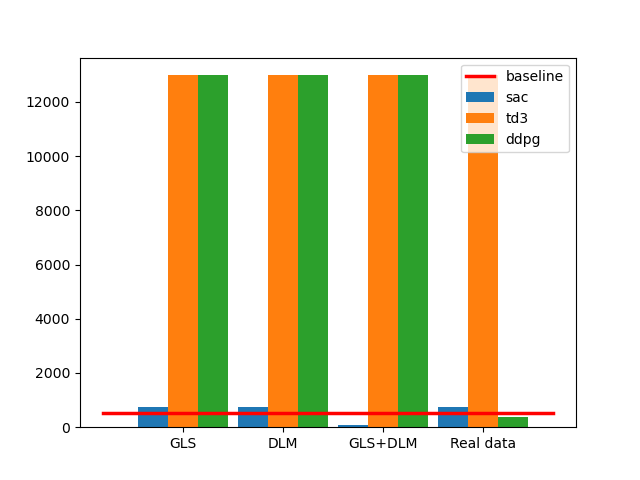}
\vspace{-10pt}
    \caption{Testing of trained policies}
    \label{fig:testing}
    \vspace{-15pt}
\end{wrapfigure}
\paragraph{Results.}  We first evaluate the different prediction model with Nash–Sutcliffe model efficiency coefficient (NSE). The results are displayed in Table~\ref{tab:nse}\footnote{We try 5 seeds and tune the algorithms' parameters, then choose the best one to plot and test.}, where we can observe that the GLS cannot provide a good prediction of the inflow from the upstream and the DLM significantly outperforms the GLS. We do not observe the improvement of including both GLS and DLM into the simulator. We present the training results in Figure~\ref{fig:training}. We observe that the training of SAC is not successful and SAC does not perform well in most of the simulators. The training of TD3 and DDPG is efficient and DDPG can significantly outperform TD3 during the training, although TD3 is more stable. We show the testing results of the trained policies, as well as the baseline strategy, in Figure~\ref{fig:testing}. We find that SAC is not perform well because the training is not successful. The policies trained by TD3 and DDPG is much better than the baseline strategy. Although DDPG performs better during the training, TD3 obtains nearly identical results during test and outperform DDPG in the simulator with real data. Interestingly, the DDPG policy trained by the simulator with the real data does not perform well when testing use the real data. One possible reason is that the policy overfits to the data used in the simulator for training, which indicates the necessity of using model for the rainfall rather than the real data in the simulator. 


\section{Conclusion}
In this work, we fist build simulators with the real data and mathematical models and use the simulator to train the efficient water release policy of the dam through deep RL. Experiments show that our methods outperform the existing baselines. Future works include building simulators with more realistic flood damage model of the downstream and more data for predicting rainfalls in the upstream and ultimately deploying to the real dams for evaluations. We target to build the AI-based dam operation system and deploy to the real dam, e.g., Bansagar dam, to improve the efficiency.

\section*{Acknowledgements}
This work is supported by NTU, ATREE and Google Research (Google grant on AI for Social Good). The ATREE team also acknowledges the support from India's National Mission on Biodiversity and Human Well-Being.


\bibliography{googledam}
\bibliographystyle{plain}

\clearpage
\appendix
\section{Details of the Baseline Discharge Strategy}
\label{app:baseline_strategy}
\begin{table}[ht]
\centering
\caption{Baseline discharge strategy from Nov to Jun. For other months, the discharge strategy will be releasing the water from the upstream, i.e., keeping the water level stable. }
\label{tab:baseline}
\begin{tabular}{ll|ll}
\toprule
Ten-Daily Period	&	Discharge (cumecs)	&	Ten-Daily Period	&	Discharge (cumecs)	\\\midrule
Nov 01-10	&	73.1	&	Mar 01-10	&	28.1	\\
Nov 11-20	&	68.6	&	Mar 11-20	&	27.0	\\
Nov 21-30	&	64.1	&	Mar 21-31	&	23.8	\\
Dec 01-10	&	59.6	&	Apr 01-10	&	20.5	\\
Dec 11-20	&	55.1	&	Apr 11-20	&	17.3	\\
Dec 21-31	&	48.5	&	Apr 21-30	&	16.1	\\
Jan 01-10	&	41.8	&	May 01-10	&	14.9	\\
Jan 11-20	&	35.2	&	May 11-20	&	13.7	\\
Jan 21-30	&	33.6	&	May 21-31	&	12.5	\\
Feb 01-10	&	31.9	&	Jun 01-10	&	11.3	\\
Feb 11-20	&	30.3	&	Jun 11-20	&	57.6	\\
Feb 21-28	&	29.2	&	Jun 21-30	&	104.0	\\
\bottomrule
\end{tabular}
\end{table}
\end{document}